\documentclass[conference]{IEEEtran}
\ifCLASSOPTIONcompsoc
  \usepackage[nocompress]{cite}
\else
  \usepackage{cite}
\fi

\ifCLASSINFOpdf
\else
\fi
\usepackage{amsmath}
\usepackage{amsfonts}
\usepackage{changebar}
\usepackage{verbatim}
\usepackage{subcaption}
\usepackage{newlfont}
\usepackage{amssymb}
\usepackage{graphicx}
\usepackage{url}
\usepackage{multicol}
\usepackage{multirow}
\usepackage{epstopdf}
\begin{document}
\setlength{\parskip}{0pt}
%
\title{$K$-Nearest Neighbor Classification Using Anatomized Data}

\author{\IEEEauthorblockN{Koray Mancuhan}
\IEEEauthorblockA{Department of Computer Science, CERIAS \\
Purdue University\\
Email: kmancuha@purdue.edu}
\and
\IEEEauthorblockN{Chris Clifton}
\IEEEauthorblockA{Department of Computer Science, CERIAS \\
Purdue University\\
Email: clifton@cs.purdue.edu}}


%


\maketitle

\begin{abstract}
	This paper analyzes $k$ nearest neighbor classification with training
	data anonymized using \emph{anatomy}. Anatomy preserves all
	data values, but introduces uncertainty in the mapping between
	identifying and sensitive values. 	We first study the theoretical effect of the 
	anatomized training data on the $k$ nearest neighbor error rate bounds,
	nearest neighbor convergence rate, and Bayesian error. We then validate 
	the derived bounds empirically. We show that 1) Learning from anatomized 
	data approaches the limits of learning through the unprotected data (although
	requiring larger training data), and 2) nearest neighbor using anatomized data 
	outperforms nearest neighbor on generalization-based anonymization.
\end{abstract}


%
\IEEEpeerreviewmaketitle

\section{Introduction}
\label{sec:intro}

Data publishing without revealing sensitive information is an 
important problem. Many privacy definitions have been proposed
based on generalizing/suppressing data ($l$-diversity\cite{ldiversity}, 
$k$-anonymity \cite{kanon_defn,kanon_sweeney}, $t$-closeness 
\cite{tcloseness}, $\delta$-presence \cite{NergizDpresence}, 
($\alpha$,$k$)-anonymity \cite{EnhancedKanon}). 
Other alternatives include value swapping \cite{CensusRelease},
distortion \cite{AgrawalPpdm}, randomization \cite{EvfimievskiBreach}, 
and noise addition (e.g., differential privacy \cite{DifferentialPrivacy}). 
Generalization consists of replacing identifying attribute values with 
a less specific version \cite{k_anon_survey}. Suppression can be 
viewed as the ultimate generalization, replacing the identifying value 
with an ``any'' value \cite{k_anon_survey}. These approaches have 
the advantage of preserving truth, but a less specific truth that 
reduces the utility of the published data.

Xiao and Tao proposed anatomization as a method to enforce 
$l$-diversity while preserving specific data values \cite{XiaoAnatomy}. 
Anatomization splits instances across two tables, one containing 
identifying information and the other containing private information.
The more general approach of fragmentation \cite{fragmentation}
divides a given dataset's attributes into two sets of attributes 
(2 partitions) such that an encryption mechanism avoids associations 
between two different small partitions. Vimercati et al. extend 
fragmentation to multiple partitions \cite{loose_fragmentation},
and Tamas et al. propose an extension that deals with multiple 
sensitive attributes \cite{l_diver_multiple}. The main advantage of
 anatomization/fragmentation is that it preserves the original
values of data; the uncertainty is only in the mapping between 
individuals and sensitive values.

We show that this additional information has real value.
First, we demonstrate that in theory, learning from anatomized data
can be as good as learning from the raw data.  We then demonstrate
empirically that learning from anatomized data beats learning from
generalization-based anonymization.

\emph{This paper looks only at instance-based learning, specifically
non-parametric $k$ nearest neighbor classifier ($k$-NN)}.  
This focus was chosen because
we have solid theoretical results on the limits 
of learning, allowing us to compare theoretical bounds on
learning from anatomized data with learning from the underlying
unprotected data.
We demonstrate this for a simple approach of
using the anatomized data; we simply consider all possible
mappings of individuals to sensitive values as equally likely.

There is concern that anatomization is vulnerable to several 
attacks \cite{Definetti,AnatomyImprov,NinghuiSlicing}.
While this can be an issue, \emph{any} method that provides
meaningful utility fails to provide perfect privacy against a
sufficiently strong adversary  \cite{NinghuiTradeoff,DifferentialPrivacy}.
Introducing uncertainty into the anonymization process reduces the
risk of many attacks,
e.g, minimality \cite{MinimalityAttack,CormodeMinimality}.
Our theoretical analysis holds for any assignment of items to
anatomy groups, including a random assignment,
which provides a high degree of robustness against
minimality and correlation-based attacks.
This paper has the following key contributions: 
\begin{enumerate}

	\item We define a classification task on anatomized data without violating the
	random worlds assumption. A violating classification task would be the
	prediction of sensitive attribute, a task that was found to be \#P-complete 
	by Kifer \cite{Definetti}.
	
	\item To our best knowledge, this is the first paper in the privacy community
	that studies the theoretical effect of training the $k$-NN on anatomized data. 
	We show the anatomization effect for the error rate bounds and the convergence 
	rate when the test data is neither anonymized nor anatomized. Inan et al. already 
	gives a practical applications of such a learning scenario \cite{kAnonSvm}.
	
	\item We show the Bayesian error estimation for any non-parametric classifier using the
	anatomized training data.

	\item We compare the $k$-NN classifier trained on the anatomized data 
	with the $k$-NN classifier trained on the unprotected data. 
	In case of nearest neighbor classifier (1-NN), we also
	make an additional comparison to generalization based learning scheme \cite{kAnonSvm}.

	\item We last compare the theoretical estimation of convergence rate with the
	practical measurements when the convergence rate is defined in function
	of $l$-diversity.
\end{enumerate}

We next summarize the related work, and give 
a set of definitions and notations necessary for further discussion. Section 
\ref{sec:errbounds} shows error rate bounds of the non-parametric $k$-NN classifier;
Section \ref{sec:bayeserr} analyzes the effect of anatomization on
the Bayesian error. Section \ref{sec:convrate} formulates the 
1-NN convergence rate under $l$-diversity. The experimental analysis is
presented in Section \ref{sec:expres}.

\section{Related Work}
\label{sec:relwork}

There have been studies in how to mine anonymized data. Nearest neighbor classification 
using generalized data was investigated by Martin. Nested generalization and non-nested 
hyperrectangles were used to generalize the data from which the nearest neighbor classifiers 
were trained \cite{MartinKNN}. Inan et al. proposed nearest neighbor and support vector 
machine classifiers using anonymized training data  that satisfy $k$-anonymity. Taylor 
approximation was used to estimate the Euclidean distance from the anonymized training 
data \cite{kAnonSvm}. Zhang et al. studied Na\"{\i}ve Bayes using partially specified training data 
\cite{Zhang2006}, proposing a conditional likehoods computation algorithm exploring the instance
space of attribute-value generalization taxonomies. 
Agrawal et al. proposed an iterative distribution reconstruction algorithm for the distorted training data
from which a C4.5 decision tree classifier was trained \cite{Agrawal01}. Iyengar suggested using a 
classification metric so as to find the optimum generalization. Then, a C4.5 decision tree classifier was 
trained from the optimally generalized training data \cite{Iyengar2002}. Fung et al.  
gave a top-down specialization method (TDS) for anonymization so that the 
anonymized data allows accurate decision trees. A new scoring function was proposed for the
calculation of decision tree splits from the compressed training data \cite{FungTDS}.
Dowd et al. studied C4.5 decision tree learning from training data perturbed by
random substitutions. A matrix based distribution reconstruction algorithm was applied
on the perturbed training data from which an accurate C4.5 decision tree classifier
was learned \cite{Dowd2006}.

None of the earlier work has provided a method directly applicable to anatomized
training data. A classifier using the anatomized training data requires specific theoretical 
and experimental analysis, because anatomized training data provides additional detail 
that has the potential to improve learning; but also additional uncertainty that must be 
dealt with. Furthermore, previous work didn't justify theoretically why the proposed heuristics
work in empirically.

\section{Definitions and Notations}
\label{sec:def}
\newtheorem{definition}{Definition}

In this section, the first four definitions will recall the standard definitions of 
unprotected data and attribute types.

\begin{definition}
	A dataset ${D}$ is called a \emph{person specific dataset} for population ${P}$ 
	if each instance ${X \in D}$ belongs to a unique individual ${p \in P}$.
\end{definition}

The person specific data will be called the training data in this paper.
Next, we will give the first type of attributes.

\begin{definition}
\label{defn:id}
	A set of attributes are called \emph{direct identifying attributes} if they let an adversary 
	associate an instance $X \in D$ to a unique individual $p \in P$ without 
	any background knowledge.
\end{definition}

\begin{definition}
\label{defn:qid}
	A set of attributes are called \emph{quasi-identifying attributes}  if there is
	background knowledge available to the adversary that associates
	the quasi-identifying attributes with a unique individual $p \in P$.
\end{definition}

We include both direct and quasi-identifying attributes under the name identifying
attribute. First name, last name and social security number (SSN) are common examples 
of direct identifying attributes.
Some common examples of quasi-identifying attributes are age, postal code,
and occupation. Next, we will give the second type of attribute.

\begin{definition}
\label{defn:s}
	An attribute of instance $X \in D$ is called a \emph{sensitive attribute} if it must be 
	protected against adversaries from correctly inferring the value for
	an individual.
\end{definition}

Patient disease and individual income are common examples of sensitive attributes. 
Unique individuals $p \in P$ typically don't want these sensitive information to be 
publicly known when a dataset $D$ is released to public. Provided an instance $X \in D$, 
the \emph{class label} is denoted by $X.C$. 
We don't consider the case where $C$ is sensitive, as this would make the purpose of classification to violate
privacy. Typically $C$ is neither sensitive nor identifying, although the analysis holds for $C$ being an
identifying attribute.

Given the former definitions, 
we will next define the anonymized training data following the definition of $k$-anonymity 
\cite{kanon_sweeney}.

\begin{definition}
	A training data $D$ that satisfies the following conditions is 
	said to be \emph{anonymized training data} $D_k$ \cite{kanon_sweeney}:
	\begin{enumerate}
		\item The training data $D_k$ does not contain any unique identifying 
		attributes.
		
		\item Every instance $X \in D_k$ is indistinguishable from at least 
		$(k-1)$ other instances in $D_k$ with respect to its quasi-identifying 
		attributes.
	\end{enumerate}
\end{definition}

In this paper, we assume that the anonymized training data $D_k$ is created according to 
a \emph{generalization} based data publishing method. We next define the 
\emph{comparison baseline classifiers}.

\begin{definition}
	A non-parametric $k$ nearest neighbor ($k$-NN) classifier that is trained on the 
	anonymized training data $D_k$ is called \emph{the anonymized $k$-NN classifier}. 
\end{definition}

\begin{definition}
	A non-parametric $k$-NN classifier that is trained on the training data $D$ is called
	\emph{the original $k$-NN classifier}. 
\end{definition}

The anonymized $k$-NN classifier will just be the comparison baseline in the evaluation 
and its theoretical discussion will not be included. We go further, requiring that there 
must be multiple possible sensitive values that could be linked to an individual. This requires
the definition of \emph{groups} \cite{ldiversity}.

\begin{definition}
\label{defn:gr}
	A \emph{group} ${G_j}$ is a subset of instances in training data ${D}$ such that 
	${D=\cup_{j=1}^{m} G_j}$, and for any pair ${(G_{j_1},G_{j_2})}$ 
	where ${1 \leq j_1 \neq j_2 \leq m}$, ${G_{j_1} \cap G_{j_2}= \emptyset }$.
\end{definition}

Next, we define the concept of $l$-diversity or $l$-diverse given the former group
definition.

\begin{definition}
\label{defn:ldiverse}
	A set of groups is said to be \emph{$l$-diverse} if and only if for all groups ${G_j}$ 
	${\forall v \in \Pi_{A_s } (G_j), \frac{freq(v,G_j )} {\vert G_j \vert} \leq \frac{1}{l}}$ where ${A_s}$ 
	is the sensitive attribute in ${D}$, $\Pi_{A_s }(*)$ is the database $A_s$ projection 
	operation on training data $*$ (or on data table in the database community), 	${freq(v,G_j )}$ 
	is the frequency of ${v}$ in ${G_j}$ and ${|G_j |}$ is the number 	of instances in ${G_j}$.
\end{definition}

We extend the data publishing method \emph{anatomization} from Xiao et al. that is originally based on 
$l$-diverse groups \cite{XiaoAnatomy}.

\begin{definition}
	\label{defn:anat}
	Given a training data ${D}$  partitioned in $m$ $l$-diverse
	groups according to Definition \ref{defn:ldiverse}, \emph{anatomization} produces an 
	\emph{identifier table} ${IT}$ and a \emph{sensitive table} ${ST}$ as follows. ${IT}$ has 
	schema
	\begin{equation*}
		(C,A_1,...,A_d,GID)
	\end{equation*}
	including the class attribute, the quasi-identifying attributes 
	${A_i \in IT}$ for ${1 \leq i \leq d}$, and 
	the \emph{group id} ${GID}$ of the group $G_j$. For each group ${G_j \in D}$ and each
	instance ${X \in G_j}$, ${IT}$ has an instance $X$ of the form:	
	\begin{equation*}
		(X.C, X.A_1,...,X.A_d,j)
	\end{equation*}
	$ST$ has schema
	\begin{equation*}
		(GID,A_s)
	\end{equation*}
	where ${A_s}$ is the sensitive attribute in ${D}$ and ${GID}$ is 
	the group id of the group $G_j$. For each group ${G_j \in D}$ and each instance 
	${X \in G_j}$, ${ST}$ has an instance of the form:
	\begin{equation*}
		(j,X.A_s)
	\end{equation*}
\end{definition}

%
	%
	%
	%
	%
%
%

Given the learning task of predicting class attribute $C$, definition \ref{defn:anat} 
lets us observe the following about training data $D$ published according to anatomization:
\emph{every instance $X_{i} \in IT$ can be matched to $l$ instances $X_j \in ST$ using 
the common attribute $GID$ in both data table schemas}. This observation yields 
the \emph{anatomized training data} and the \emph{anatomized $k$-NN classifier}.

\begin{definition}
	\label{def:anatr}
	Given two data tables $IT$ and $ST$ resulting from the anatomization on training data $D$, 
	the \emph{anatomized training data} $D_A$ is 
	%
	\begin{equation*}
		D_A=\Pi_{IT.A_1, \cdots IT.A_d,ST.A_s} ( \; IT \Join \; ST)
	\end{equation*}
	where $\Join$ is the database inner join operation with respect to the
	condition $IT.GID = ST.GID$ 
	and $\Pi(*)$ is 
	the database projection operation on training data (*) processed according to definition 
	\ref{defn:anat}.
\end{definition}

\begin{definition}
	A non-parametric $k$-NN classifier that is trained 
	on the anatomized training data $D_A$ is called \emph{the anatomized k-NN classifier}.
\end{definition}

Using the former definitions, we now give assumptions and notations
used in discussing the anatomized $k$-NN classifier.
In the theoretical analysis, we assume that all the training data has a smooth probability distribution.
Although anatomization requires a discrete probability distribution for the sensitive attribute 
$A_s$, such smoothness violation is negligible since the original $k$-NN classifier is known to fit well on discrete 
training data \cite{tanBook}. The sensitive attribute $A_s$ is assumed to be non-binary.
The anatomized $k$-NN cases where $k>1$ and $k$ is even will be ignored,
because such cases include the tie between $k$-nearest neighbors that makes the bounds 
ambiguous and complicated \cite{fukunagaBook}. The total number of attributes are assumed to be $d+1$ 
($d$ identifying attributes and 1 sensitive attribute) and all instances are assumed to be in a separable 
metric space $M \subset \mathbb{R}^{d+1}$
as in \cite{Cover67,devroyeBook,fukunagaBook}. 
$D$ has $N$ instances whereas $D_A$ has $Nl$ instances from definition \ref{def:anatr}.
All instances are i.i.d whether they are
in training or test data.
For the sake of simplicity, $A_{id}$ 
will denote the identifying attributes $A_1 \cdots A_d \in IT$. $T$ stands for a test 
data which is not processed by any anatomization and generalization method.
$X$ will be an instance of the test data $T$. 
$d(U, V)$ is the quadratic distance metric for a pair of instances $U$ and $V$ in metric space $M$. 
$X^\prime_N(k)$ denotes the set of $k$ number of nearest neighbors of 
$X$ in $D$ that the original $k$-NN classifier uses while $X^\prime_{Nl}(k)$ denotes the set of $k$ number 
of nearest neighbors of $X$ in $D_A$ that the anatomized $k$-NN classifier uses.
$X_i$ will interchangeably be an instance of $D$ or $D_A$ and $X_j$ will interchangeably be an instance of $X^\prime_N(k)$
or $X^\prime_{Nl}(k)$. 
In case of $k=1$, we will use $X^\prime_N$ and $X^\prime_{Nl}$ for the nearest neighbors in $D$ and $D_A$.
$\mathbf{X}$ is the random variable with probability distribution $P(\mathbf{X})$ from which $X$ and $X_i$ are drawn.
Training and test instances will be column vectors in format of $(A_1,...,A_d,A_s)^T$.
$C$ is the class attribute in $D$ and $D_A$ with binary labels 1 and 2. 
Given the training data $D$ and the class label $i$, $q_i(X)$, 
$P_i(X)$ and $P_i$ stand for the posterior probability, the likelihood probability and the 
prior probability respectively. If the anatomized training data $D_A$ is used, 
 $q_{A_i}(X)$, $P_{A_i}(X)$ and $P_{A_i}$ are the symmetric definitions for the class label $i$.
$R(X_N^\prime(k), X)$ is the error rate when $X \in T$ is classified using $X_N^\prime(k)$. 
If $X_{Nl}^\prime(k)$ is used to classify $X$, $R_A(X_{Nl}^\prime(k), X)$ will be the error rate.
When $X_j \cong X$ hold for all $X_j \in X_N^\prime(k)$, we denote the error rate 
by $R^k(X)$ in Equation \ref{eq:errX} \cite{fukunagaBook}.
\begin{equation}
	\label{eq:errX}
	\begin{split}
		R^k(X) &= \overset{k+1/2}{\underset{i=1}{\sum}} \frac{1}{i} \binom{2i-2}{i-1}[q_1(X)q_2(X)]^i \\
		&+ \frac{1}{2} \binom{k+1}{k+1/2} [q_1(X)q_2(X)]^{k+1/2}
	\end{split}
\end{equation}
$R_A^k(X)$ is the error rate when $X_j \cong X$ hold for all $X_j \in X_{Nl}^\prime(k)$. 
$R_A^k(X)$ can trivially be derived from Eqn. \ref{eq:errX} by substituting $q_i(X)$ with $q_{A_i}(X)$.
The Bayesian errors given $X$ are denoted by $R^*(X)$ and $R_A^*(X)$ when $X_j \cong X$ holds 
for all $X_j \in X_{N}^\prime(k)$ and $X_j \in X_{Nl}^\prime(k)$ respectively.
Eqn. \ref{eq:bayeserrX} computes $R^*(X)$ \cite{fukunagaBook}.
\begin{align}
	\label{eq:bayeserrX}
	\begin{split}
		R^*(X) &= min \{ q_1(X), q_2(X)\} \\
		&\cong\overset{\infty}{\underset{i=1}{\sum}} \frac{1}{i} \binom{2i-2}{i-1}[q_1(X)q_2(X)]^i 
	\end{split}
\end{align}
$R_A^*(X)$ can trivially be derived again from \ref{eq:bayeserrX} by substituting $q_i(X)$ with $q_{A_i}(X)$.
$R^k$ and $R_A^k$, which are $E\{ R^k(X) \}$ and $E\{ R_A^k(X) \}$ with respect to $\mathbf{X}$, will 
stand for the error rate of original $k$-NN and anatomized $k$-NN classifiers respectively. $R^*$ and $R_A^*$, 
which are $E\{ R^*(X) \}$ and $E\{ R_A^*(X) \}$ with respect to $\mathbf{X}$, will stand for the Bayesian 
errors of original training data and anatomized training data respectively. We will denote $R^1(X)$ and 
$R^1_A(X)$ by $R(X)$ and $R_A(X)$ for convenience. Similarly, $R$ and $R_A$ will denote $R^1$ and 
$R^1_A$. Further notations and definitions will be given in the paper if necessary. 

\section{Error Bounds of Anatomized $k$-NN}
\label{sec:errbounds}
\newtheorem{corollary}{Corollary}
\newtheorem{theorem}{Theorem}
\newtheorem{proofsk}{Proof Sketch}

In this section, we will first show the error bounds for the anatomized 1-NN classifier. 
We will then discuss the extension to the anatomized $k$-NN classifier for all odd $k>1$. 
We give only proof sketches due to space limitations.

We first give Corollary \ref{cor:1-NNconvana} which is critical for the error bounds of the anatomized
1-NN classifier.

\begin{corollary}
	\label{cor:1-NNconvana} 
	\emph{Convergence of the nearest neighbor in the anatomized training data $D_A$.}
	Let $X \in T$ and ${X}_1, \cdots, X_{Nl} \in D_A$ be i.i.d instances taking values separable
	in any metric space $M \subset \mathbb{R}^{d+1}$. Let $X^\prime_{Nl}$ be the nearest 
	neighbor of $X$ in $D_A$. Then, $\underset{N \to \infty} {\lim} X_{Nl}^\prime = X$ with 
	probability one.
\end{corollary}

We can intuitively say that Corollary \ref{cor:1-NNconvana} should hold for the anatomized
training data $D_A$ if it already holds for the training data $D$. For the nearest neighbor 
$X^\prime_N \in D$ of $X$, there are $l$ instances in the anatomized training data $D_A$ including 
$X^\prime_N$ itself. Assuming very large training data size ($N \to \infty$), $X^\prime_N$ 
must still be the closest instance to $X$ in the anatomized training data $D_A$. The $l-1$ incorrect 
instances are expected to remain far and $X^\prime_{Nl}=X^\prime_N$ should eventually hold.

We now give a sketch of the proof or Corollary \ref{cor:1-NNconvana}. Let 
$S_{X}(r)=\{ \bar{X} \in M: d(X,\bar{X}) \leq r\}$ be the 
sphere with radius $r>0$ centered at $X$.  Let's consider that $X$ has a sphere $S_{X}(r)$ with 
non-zero probability. Therefore, for any radius $\delta>0$ and any fixed $l \geq 0$;
\begin{equation}
	\label{eq:1-NNconvana}
	\begin{split}
		P\{ \min_{i=1,\cdots,Nl} d({X}_i,X) &\geq \delta \}=[1-P(S_X(\delta))]^{Nl} \\
		&\cong \underset{N \to \infty} {\lim} [(1-P(S_X(\delta)))^{l}]^N \\
		&= 0
	\end{split}
\end{equation}
Since $d(X_i,X)$ is monotonically decreasing in terms of $i$ for all $X_i \in D_A$, we can conclude that 
$\underset{N \to \infty} {\lim} X_{Nl}^\prime = X$ holds with probability 1. The rest of proof follows
the denseness of the set $\mathbb{Q}$ in the set $\mathbb{R}$ according to Cover et al. \cite{Cover67}.

Next, Theorem \ref{th:1-NNboundsana} shows the error bounds of the anatomized 1-NN classifier
using Corollary \ref{cor:1-NNconvana}.

\begin{theorem}
	\label{th:1-NNboundsana}
	\emph{Error Rate Bounds of the anatomized 1-NN classifier}
	Let $M \subset \mathbb{R}^{d+1}$ be a metric space. Let $P_{A_1}(X)$ and $P_{A_2}(X)$ 
	be the likelihood probabilities of $X$ such that $P_A(X)=P_{A_1} P_{A_1}(X)+P_{A_2} P_{A_2}(X)$ 
	with class priors $P_{A_1}$ and $P_{A_2}$. Last, let's assume that $X$ is either a point of non-zero 
	probability measure or a continuity point of $P_{A_1}(X)$ or $P_{A_2}(X)$. Then the nearest neighbor 
	has the probability of error $R_A$ with the bounds
	\begin{equation}
		\label{eq:bounds}
		R^*_A \leq R_A \leq 2 R^*_A
	\end{equation}
	where $R^*_A$ denotes the Bayesian error when the anatomized training data $D_A$ is used.
\end{theorem}

We now give a sketch of proof for Theorem \ref{th:1-NNboundsana}.
Let $R_A(X^\prime_{Nl}, X)$ denote the probability of error for a pair of instances $X \in T$ and 
$X^\prime_{Nl} \in D_A$.
Since Corollary \ref{cor:1-NNconvana} shows that $\underset{N \to \infty} {\lim} X_{Nl}^\prime = X$ 
always hold,	\ref{eq:anerrorrate} is derived from \ref{eq:errX} by substituting $k$ with 1 and $q_i(X)$ 
with $q_{A_i}(X)$.
\begin{equation}
	\label{eq:anerrorrate}
	\underset{N \to \infty} {\lim} R_A(X^\prime_{Nl}, X)=R_A(X)=2q_{A_1}(X)q_{A_2}(X)
\end{equation}
The rest of the derivation follows Cover et al.  using \ref{eq:errX}, \ref{eq:bayeserrX} \cite{Cover67}. 

Extending \ref{eq:bounds} from the anatomized 1-NN classifier to the anatomized $k$-NN classifier for all odd
$k>1$ follows the steps in Corollary \ref{cor:1-NNconvana} and Theorem \ref{th:1-NNboundsana}.
The key is to show that $\underset{N \to \infty} {\lim} X_j= X$ holds for all $X_j \in X^\prime_{Nl}(k)$. 
The rest is to derive an expression of $R_A^k(X)$ as in \ref{eq:anerrorrate} 
for all odd $k>1$ and show that $R_A^k(X)$ is always less than $2 R_A^*$ and $R_A^{k-2}(X)$. We exclude this 
derivation due to space limitations, but the derivation follows from the original $k$-NN classifier analysis in
\cite{fukunagaBook}.
The anatomized $k$-NN classifier has the bound \ref{eq:boundsk}
\begin{equation}
	\label{eq:boundsk}
	R_A^* \leq \cdots \leq R_A^5 \leq R_A^3 \leq R_A \leq 2 R_A^*
\end{equation}
for all odd $k>1$.

Note that the Bayesian errors $R^*_A$ and $R^*$ are not always same 
due to the $l$-diverse groups of the anatomization. The $l$-diverse groups cause new likelihood $P_{A_i}(X)$
and eventually posterior probabilities $q_{A_i}(X)$. $R^*_A$ thus differ from \ref{eq:bayeserrX}, because
\ref{eq:bayeserrX} uses $q_i(X)$ instead of $q_{A_i}(X)$. The next section formulates this change.

\section{Bayesian Error On Anatomized training data}
\label{sec:bayeserr}
\newtheorem{lemma}{Lemma}
\newtheorem{axiom}{Axiom}

Since it is impossible to know the exact Bayesian error, many Bayesian error estimation techniques 
were suggested \cite{devroyeBook,fukunagaBook, devroye99}. 
In this section, the Bayesian error will be estimated for binary classification using Parzen density estimation. 
Although such estimation 
would be very interesting for multi-label classification, the theoretical analysis on unprotected data only covers
binary classification \cite{devroye99}.
The Parzen density estimation approach, which is easier to derive than the $k$ nearest neighbor density estimation 
approach, will follow Fukunaga \cite{fukunagaBook} and Fukunaga et al. \cite{fukunagaBayes87}.
Both approaches show the same behavior in terms of the Bayesian estimation that makes the discussion general enough 
for any non-parametric density based binary classification method \cite{fukunagaBook}. We first give three 
axioms and a lemma.
\begin{axiom}
	\label{ax:bayesian1}
	Given the anatomized training data $D_A$ and the training data $D$; let $P_i$ and $P_{A_i}$ be 
	the class priors for class labels $i=\{1, 2\}$. Then, $P_{i}=P_{A_i}$ is always true.
\end{axiom}
	\begin{axiom}
		\label{ax:bayesian2}
		Let $P_{1} P_{1}(X.A_{id})+P_{2} P_{2}(X.A_{id})$ and 
		$P_{A_1} P_{A_1}(X.A_{id})+P_{A_2} P_{A_2}(X.A_{id})$ 
		be $P(X.A_{id})$ and $P_A(X.A_{id})$ respectively.
		Given the anatomized training data $D_A$ and the training data $D$; let $P(X.A_{id})$ and 
		$P_{A}(X.A_{id})$ be the smooth joint densities of identifying attributes $A_{id}$. 
		Then, $P(X.A_{id})=P_{A}(X.A_{id})$ is always true.
	\end{axiom}
	\begin{axiom}
		\label{ax:bayesian3}
		Let $P_{1} P_{1}(X.A_{s})+P_{2} P_{2}(X.A_{s})$ and 
		$P_{A_1} P_{A_1}(X.A_{s})+P_{A_2} P_{A_2}(X.A_{s})$ 
		be $P(X.A_{s})$ and $P_A(X.A_{s})$ respectively.
		Given the anatomized training data $D_A$ and the training data $D$; let $P(X.A_s)$ and 
		$P_{A}(X.A_s)$ be the smooth densities of sensitive attribute $A_s$. 
		Then, $P(X.A_{s})=P_{A}(X.A_{s})$ is always true.
	\end{axiom}
Axioms \ref{ax:bayesian1}, \ref{ax:bayesian2} and \ref{ax:bayesian3} are obvious due to
the following: \emph{provided a sample of size N drawn from a probability distribution $P$, repeating
every instance for fixed $l>0$ times and obtaining a sample of size $Nl$ does not change the
probability distribution $P$. The estimated parameters $\widehat{\mu}$ and $\widehat{\sigma^2}$ 
of distribution $P$ remain same}.
\begin{lemma}
	\label{lem:bayesian1}
	Given the anatomized training data $D_A$ and the training data $D$, let identifying attributes 
	$A_{id}$ and the sensitive attribute $A_{s}$ be independent. Then, $P_A(X)=P(X)$ is always 
	true under the axioms \ref{ax:bayesian2} and \ref{ax:bayesian3}.
\end{lemma}
Using axioms \ref{ax:bayesian2} and \ref{ax:bayesian3}, the
proof of lemma \ref{lem:bayesian1} is straightforward. Lemma \ref{lem:bayesian1} 
and axioms 1-3 yield the Theorem \ref{th:bayesianest}. Using lemma \ref{lem:bayesian1}, 
we will assume that $R^*_A=R^*$ holds asymptotically for Bayesian errors.
\begin{theorem}
	\label{th:bayesianest}
	Let $M \subset \mathbb{R}^{d+1}$ be a metric space.
	Let $P_{A_1}(X)$ and $P_{A_2}(X)$ be the smooth probability density 
	functions of $X$. Let $P_{A_1}$ and $P_{A_2}$ be the class priors  such 
	that $P_A(X)=P_{A_1} P_{A_1}(X)+P_{A_2} P_{A_2}(X)$.  Similarly, 
	let $P_{1}(X)$ and $P_{2}(X)$ be the  smooth probability density functions
	of $X$ such that $P(X)=P_{1} P_{1}(X)+P_{2} P_{2}(X)$ with class 
	priors $P_{1}$ and $P_{2}$. Let $h_A(X)=-ln(\frac{P_{A_1}(X)}{P_{A_2}(X)})$ and 
	$h(X)=-ln(\frac{P_1(X)}{P_2(X)})$ be the classifiers with 
	biases $\Delta h_A(X)$ and $\Delta h(X)$ respectively. Let 
	$t=ln(\frac{P_{A_1}}{P_{A_2}})=ln(\frac{P_{1}}{P_{2}})$ 
	be the decision threshold with threshold bias $\Delta t$. Let $\epsilon_A >0$ 
	be the small changes on $P_{1}(X)$ and $P_{2}(X)$ resulting in $P_{A_1}(X)$ 
	and $P_{A_2}(X)$; and $\widehat{R}^*_A$, $\widehat{R}^*$ be the Bayesian error estimations 
	with respective biases $\Delta R^*_A$, $\Delta R^*$. Let $\widehat{P}_{A_i}(X)$ and $\widehat{P}_i(X)$ be 
	the Parzen density estimations; and $K(*)$ be the kernel function for $D$ with shape matrix 
	$A$ and size/volume parameter $r$ \cite{fukunagaBook}.
	Last, let's assume that 1)$A_{id}$ and $A_s$ are independent in the training data $D$ 
	and the anatomized training data $D_A$ 2) $R^*_A=R^*$ hold 3) $\Delta t<1$.
	Therefore,
	\begin{equation}
		\label{eq:bayesdecana}
		\begin{split}
		\widehat{R}^*_A \cong R^*+a_1 r^2 &+ a_2 r^4 + a_3 \frac{r^{-(d+1)}}{N} \\
		&+ \epsilon_A a_4 r^2+ \epsilon_A a_5 r^4  - \epsilon_A a_6 \frac{r^{-(d+1)}}{N}
		\end{split}
	\end{equation}
	where $\epsilon_A a_6 \frac{r^{-(d+1)}}{N} >0$ always holds.
\end{theorem}
Due to lack of space, we provide a brief summary of the proof. In \ref{eq:bayesdecana},
the terms other than $R^*$ stand for the expected estimation error $E[\Delta R^*_A]$ in 
\ref{eq:bayeserrbias} \cite{fukunagaBook}.
\begin{equation}
	\label{eq:bayeserrbias}
	\begin{split}
	E[\Delta R_A^*] \cong& \\
	\frac{1}{2\pi} \int \int &E[\Delta h_A(X) + 
	\frac{(j\omega)}{2} \Delta h^2_A(X)] e^{j\omega h_A(X)} \\ 
	&\times [P_{A_1} P_{A_1}(X) - P_{A_2} P_{A_2}(X)] d\omega dX
	\end{split}
\end{equation}
Hence, the proof of this theorem requires the second order approximations of $E\{\Delta h_A(X)\}$ 
and $E\{\Delta h^2_A(X)\}$. From Fukunaga \cite{fukunagaBook}, we know that $E\{\Delta h_A(X)\}$ and $E\{\Delta h^2_A(X)\}$
are expressed in function of the $E\{ \widehat{ P}_{A_i}(X) \} $ and $Var\{ \widehat{P}_{A_i}(X)\}$.
The key point of the proof is to formulate the anatomized training data effect in $E\{ \widehat{ P}_{A_i}(X) \} $
and $Var\{ \widehat{P}_{A_i}(X)\}$ and show its propagation to the $E\{\Delta h_A(X)\}$ and 
$E\{\Delta h^2_A(X)\}$. Let $\epsilon_A>0$ be the small change in the likelihood probabilities 
$P_i(X)$ which results in $P_{A_i}(X)$, $t$ be $ln(P_1/P_2)$ and $t=t_A$ be true due to axiom \ref{ax:bayesian1}. 
Therefore, we have \ref{eq:like1} and \ref{eq:like2} as the likelihood densities in the anatomized training data $D_A$
using lemma \ref{lem:bayesian1}.
\begin{align}
	\label{eq:like1}
	P_{A_1}(X) &= P_1(X) + \epsilon_A \\
	\label{eq:like2}
	P_{A_2}(X) &= P_2(X) - e^t \epsilon_A
\end{align}
Using \ref{eq:like1} and \ref{eq:like2} in the Taylor approximations of $E\{ \widehat{ P}_{A_i}(X) \} $
and $Var\{ \widehat{P}_{A_i}(X)\}$ results in the approximations of $E\{\Delta h_A(X)\}$  in \ref{eq:hbiasana}
\begin{equation}
	\label{eq:hbiasana}
	\begin{split}
		E\{\Delta h_A(X)\} &\cong E\{ \Delta h(X) \}\\
		&+ \epsilon_A \frac{r^2}{2}[\frac{\alpha_1(X)}{P_1(X)}+ e^t\frac{\alpha_2(X)}{P_2(X)}]\\
		&- \epsilon_A \frac{r^4}{4} [\frac{\alpha_1^2(X)}{P_1(X)} + e^t \frac{\alpha_2^2(X)}{P_2(X)}]\\
		&-\epsilon_A \frac{r^{-(d+1)}}{2N} [\frac{s_1}{P_1^2(X)} + e^t \frac{s_2}{P_2^2(X)}]
	\end{split}
\end{equation}
and $E\{\Delta h^2_A(X)\}$ in \ref{eq:hbiasanasq}
\begin{table*}[htb!]
\centering
\caption{Summary of Theoretical Analysis}
\label{tab:summary}
\begin{tabular}{|c|c|c|c|c|}
\hline
& \textbf{Training Data $D$} & \textbf{Anatomized Training Data $D_A$} & \textbf{Notations}  \\ \hline
\multirow{6}{*}{\textbf{$k$-NN Error Rate Bounds}} 
	& \multirow{6}{*}{ $R^* \leq \cdots \leq R^5 \leq R^3 \leq R \leq 2 R^*$} &
      \multirow{6}{*}{$R_A^* \leq \cdots \leq R_A^5 \leq R_A^3 \leq R_A \leq 2 R_A^*$ }
      &$R$: 1-NN error rate ($D$) \\
      & & & $R^k$: $k$-NN error rate ($D$) \\
    & & & $R^*$: Bayesian error ($D$)\\
    & & &$R_A$: 1-NN error rate ($D_A$) \\
    & & & $R_A^k$: $k$-NN error rate ($D_A$)\\
    & & &  $R^*_A$: Bayesian error ($D_A$) \\
\hline
\multirow{3}{*}{\textbf{1-NN Convergence Rate}} & \multirow{3}{*}{ $O(1/N^{2/d+1})$} &
      \multirow{3}{*}{$O(1/(Nl)^{2/d+1})$ }
      &$N$: Number of training instances\\
    & & &  $l$: $l$-diversity parameter\\
    & & &$d$: Number of identifying attributes \\
    \hline
\multirow{6}{*}{\textbf{Bayesian Error Estimation}} & \multirow{6}{*}{ $R^*+a_1 r^2 + a_2 r^4 + a_3 \frac{r^{-(d+1)}}{N}$} &
      \multirow{6}{*}{$\widehat{R}^* +\epsilon_A a_4 r^2+ \epsilon_A a_5 r^4  - \epsilon_A a_6 \frac{r^{-(d+1)}}{N}$ }
      & $R^*$: Bayesian error ($D$) \\
    & & &  $\widehat{R}^*$: Bayesian error estimation for $D$\\
    & & & $r$: Kernel width parameter \\
    & & & $N$: Number of training instances \\
    & & & $\epsilon_A$: Small change on likelihood \\
    & & & $d$: Number of identifying attributes \\
    \hline
\end{tabular}
\end{table*}
\begin{equation}
	\label{eq:hbiasanasq}
	\begin{split}
	E\{\Delta &h^2_A(X)\} \cong E\{\Delta h^2(X)\} \\
	&- \epsilon_A \, \Delta t \, r^2 [ \frac{\alpha_1(X)}{P_1(X)} + e^t\frac{\alpha_2(X)}{P_2(X)}] \\
	&+ \epsilon_A \frac{r^4}{2} [\frac{\alpha_1(X)\alpha_2(X)}{P_1(X)} -
	e^t \frac{\alpha_1(X) \alpha_2(X)}{P_2(X)}] \\
	&- \epsilon_A \frac{r^4}{2} [\frac{\alpha_1^2(X) (1-\Delta t) }{P_1(X)} - 
	e^t \frac{\alpha_2^2(X) (1+\Delta t)}{P_2(X)}] \\
	&- \epsilon_A \frac{r^{-(d+1)}}{N} [\frac{(1-\Delta t)s_1}{P_1^2(X)} + e^t \frac{(1+\Delta t) s_2}{P_2^2(X)}]
	\end{split}
\end{equation}
where $w_i=s_i r^{-(d+1)}$ is true. The former equality is the result of using Parzen 
density estimate \cite{fukunagaBook}. \ref{eq:hbiasana} and \ref{eq:hbiasanasq} are
derived using the Taylor approximations up to second order. Plugging \ref{eq:hbiasana} 
and \ref{eq:hbiasanasq} in \ref{eq:bayeserrbias} and rewriting \ref{eq:bayeserrbias} 
gives \ref{eq:bayesdecana} where each $a_i$ stands for an integration term.

Eqn. \ref{eq:bayesdecana} shows that the anatomized training data $D_A$ reduces the variance
term of the decision functions that estimate the Bayesian error. However, it is hard 
to determine the effect of the anatomized training data $D_A$ on bias terms. All 
$\epsilon_A a_4 r^2>0$, $\epsilon_A a_4 r^2<0$, $ \epsilon_A a_5 r^4>0$ and 
$ \epsilon_A a_5 r^4<0$ are possible cases depending on $h_A(X)$ which might 
yield bias terms of $\widehat{R}^*_A$ bigger or smaller than $\widehat{R}^*$'s ones. 

\section{Anatomized 1-NN Convergence}
\label{sec:convrate}

We now discuss the error rate of the anatomized 1-NN classifier
when the anatomized training data $D_A$ has finite size $Nl$. We will then derive 
the convergence rate from the former error rate. The discussion here won't be generalized
to the anatomized $k$-NN classifier since the finite size training data performance of
$k$-NN classifiers are not generalized to $k>2$ in the pattern recognition literature 
\cite{devroyeBook, fukunagaBook}. Also, only binary classification will be considered
due to space limitations. 

From Theorem \ref{th:bayesianest}, we 
intuitively expect a faster convergence rate than the
original 1-NN classifier's one. For $N$ number of instances in training data $D$, using the 
anatomized training data $D_A$ reduces the variance of any classifier's Bayesian error estimation. Therefore, there 
are fewer possible models to consider for a given sample size which eventually means a faster convergence 
to the asymptotic result. Theorem \ref{th:finiteerrana} extends the analysis of Fukunaga et al.
\cite{fukunagaBook,fukunagaBias87}. 

\begin{theorem}
	\label{th:finiteerrana}
	Let $M \subset \mathbb{R}^{d+1}$ be a metric space.
	Let $P_{A_1}(X)$ and $P_{A_2}(X)$ be the smooth probability density 
	functions of $X$. Let $P_{A_1}$ and $P_{A_2}$ be the class priors  such 
	that $P_A(X)=P_{A_1} P_{A_1}(X)+P_{A_2} P_{A_2}(X)$.	
	Let $q_{A_1}(X)$ and $q_{A_2}(X)$ be the smooth posterior probability densities such 
	that $q_{A_1}(X)+q_{A_2}(X)=1$ and $Nl \to \infty$. Let $q_{A_1}(X^\prime_{Nl})$ and $q_{A_2}
	(X^\prime_{Nl})$ be the smooth posterior probability densities  such that 
	$q_{A_1}(X^\prime_{Nl})+q_{A_2}(X^\prime_{Nl})=1$ and $Nl \nrightarrow \infty$. 
	Let $\delta >0$ be the difference between $q_{A_i}(X)$ and $q_{A_i}(X^\prime_{Nl})$ for 
	class labels $i=\{ 1, 2 \}$.  Let $d(X^\prime_{Nl}, X)$ be the quadratic distance with matrix $A$ and
	$\rho$ be the calculated value of $d(X^\prime_{Nl}, X)$.
	Let $R_A$ be the error rate of the anatomized 1-NN classifier when $Nl \to \infty$. Last, 
	let $R_{A_N}$ be the error rate of the anatomized 1-NN classifier when $Nl \nrightarrow \infty$. Then,
	\begin{equation}
		\label{eq:errorratefinite}
		R_{A_N}\cong R_A
		+\beta \frac{1}{(Nl)^\frac{2}{d+1}} E_X\{ \vert A \vert ^{-\frac{1}{d+1}} tr\{ A B(X)\} \}
	\end{equation}
	where $\beta$ is
	\begin{equation}
		\beta=\frac{\Gamma^\frac{2}{d+1}(\frac{d+3}{2}) \Gamma(\frac{2}{d+1}+1)}{\pi (d+1) }
	\end{equation}
	and $B(X)$ is
	\begin{equation}
		\begin{split}
		B(X)&=P_A^{-\frac{2}{d+1}}(X) [q_{A_2}(X) -q_{A_1}(X)] \\
		&\times [ \frac{1}{2} \nabla^2 q_{A_1}(X) + 	P_A^{-1}(X) \nabla P_A(X) \nabla^T q_{A_1}(X) ]
		\end{split}
	\end{equation}
\end{theorem}

We will give here a summary of proof. We first define $q_{A_i}(X^\prime_{Nl})$ in function 
of $q_{A_i}(X) \pm \delta$ such that $q_{A_1}(X^\prime_{Nl})+q_{A_2}(X^\prime_{Nl})=1$ holds. 
Then, $R_{A_N}$ is written in function of $R_A$ and $\delta$. The result is
\begin{equation}
	\label{eq:avgerrrate}
	R_{A_N}=R_A+E[[q_{A_2}(X)-q_{A_1}(X)]\delta]
\end{equation}
where $E[[q_{A_2}(X)-q_{A_1}(X)]\delta]$ is
\begin{equation}
	\label{eq:bias}
	\begin{split}
	E \{ (q_{A_2}(X)-q_{A_1}(&X)) \delta) \}= \\
	E_X\{ E_\rho \{ E_{X^\prime_{Nl}}\{ &[q_{A_2}(X) - q_{A_1} (X)] \delta \vert \rho, X \} \vert  X \}  \}
	\end{split}
\end{equation}
a 3-step expectation in \ref{eq:bias}. The rest of the proof follows Fukunaga \cite{fukunagaBook}.
The key deviation of the anatomized training data $D_A$ from the training data $D$ results from
the step 2. In step 2, the nearest neighbor density estimation is done on $Nl$ training
instances instead of $N$ training instances. Thus, the expectation with respect to $\rho$ gives
\ref{eq:exprho3}.
\begin{equation}
	\label{eq:exprho3}
	E \{\rho^2\} \cong \frac{\Gamma^\frac{2}{d+1}(\frac{d+3}{2}) \Gamma(\frac{2}{d+1}+1)}
	{P_A^\frac{2}{d+1}(X) \pi \vert A \vert ^\frac{1}{d+1}} \frac{1}{(Nl)^\frac{2}{d+1}}	
\end{equation}
Using \ref{eq:exprho3}, expectation with respect to $X$ in \ref{eq:bias} (step 3) according to Fukunaga 
\cite{fukunagaBook} results in \ref{eq:errorratefinite}. Table \ref{tab:summary} gives a summary of 
theoretical analysis, including a comparison between the anatomized training data $D_A$ and the training 
data $D$.

\section{Experiments and Results}
\label{sec:expres}

\subsection{Preprocessing, Setup and Implementation}

We evaluate the anatomized $k$-NN classifier using cross validation on the \emph{Adult}, \emph{Bank Marketing}, 
\emph{IPUMS} datasets from UCI collection \cite{UCIrepository} and on the \emph{Fatality (fars)} dataset 
from Keel repository \cite{keel10}.

In the adult dataset, we predicted the income attribute. The instances with missing values were
removed and features selected using the Pearson correlation filter (CfsSubsetEval) of Weka \cite{Weka}.
After preprocessing, we had 45222 instances with 5 attributes education, marital status, 
capital gain, capital loss and hours per week and the class attribute income.
The other datasets were used without feature selection. In IPUMS, 
we predicted whether a person is veteran or not. After removing
the N/A and missing values for veteran information, there were 148585 instances 
with 59 attributes. In the Fatality dataset, we predicted whether a person is injured or not in a car accident
based on 29 attributes. Since the class attribute was non-binary in the original data, the instances
with class labels ``Injured\_Severity\_Unknown'', ``Died\_Prior\_to\_Accident'' and ``Unknown'' were removed
and the binary class values ``Injured'' vs ``Not\_Injured'' were created. The former removal resulted in 
91085 instances. In the Bank Marketing dataset, we predicted whether a person replied positively or negatively to 
the bank's phone marketing campaign. The dataset is used with 41188 instances and 20 attributes.

In the Adult, Bank Marketing and IPUMS datasets, education (educrec in IPUMS) was deemed sensitive whereas the 
remaining attributes were quasi-identifying attributes. Education had many discrete values which
lets all samples satisfy $l$-diversity when $l=2,3$. In the Fatality dataset, ``POLICE\_REPORTED\_ALCOHOL\_INVOLVEMENT''
was the sensitive attribute whereas rest of the attributes were quasi-identifying attributes. This was the
only discrete attribute in the dataset other than class attribute that is not a typical quasi-identifying attribute
such as state, age, zipcode.

Weka (same version of Inan et al. \cite{kAnonSvm}) was used to implement the $k$-NN classifier \cite{Weka}. The 
anatomization algorithm was implemented by us following Xiao et al. \cite{XiaoAnatomy}. 
All the anatomized training data were created from identifier and sensitive tables using the merge 
function of R. The error rates were measured on each test fold according to the definition in 
Weka implementation. 
When we compared the anatomized 1-NN with anonymized 1-NN, we also used 
the same generalization hierarchies that Inan et al. used. 
The statistical tests following Kumar et al. are provided \cite{tanBook}

\begin{figure}[htb!]
                    \centering
                    \includegraphics[width=.35\textwidth,height=0.35\textwidth]{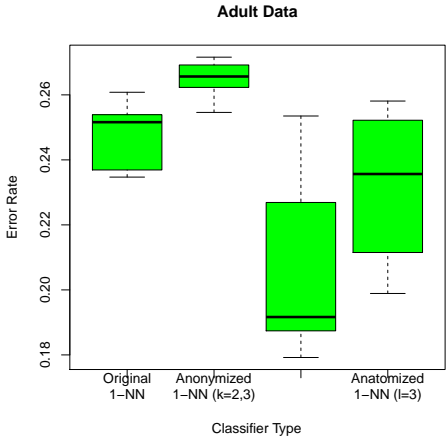}
                    \caption{Error Rate on 10 Fold Cross Validation} 
                    \label{fig:err10}
\end{figure}

\subsection{Anatomized 1-NN vs Anonymized 1-NN and Original 1-NN}

First, we compare the anatomized 1-NN classifier with both anonymized 
and original 1-NN classifiers. We consider anonymized and anatomized training data with 
the quasi-identifying groups having similar number of instances ($k=l$). 
Figure \ref{fig:err10} shows the plot of error rates on 10-fold cross validation without outlier values.
We give results for two scenarios: 1) $k=l=2$ vs original data 
2) $k=l=3$ vs original data. 
Although we measured the error rates to $k=l=7$, we omit these results due to space limitations. 
The results are similar when $k=l>4$ even though some instances are suppressed to maintain 
$l$-diversity.

In Figure \ref{fig:err10}, the general trend is that anatomized 1-NN has the smallest error rates and 
anonymized 1-NN has the largest error rates. 
The average error rates for anonymized 1-NN and anatomized 1-NN classifiers are 0.3132 and 0.204 for
$k=l=2$ and 0.3132 and 0.2324 for $k=l=3$.
Meanwhile, the original 1-NN 
has average error rate of 0.2456. When $k=l=2$, the anatomized 1-NN has significantly lower error rates 
than the original 1-NN 
at the confidence intervals 0.99, 0.98, 0.95, 0.9 and 0.8. When $k=l=3$, the anatomized 1-NN has significantly lower error 
rates than the original 1-NN at the confidence interval 0.99.
This is a surprising and an interesting result showing
the practical interpretation of Theorem \ref{th:bayesianest} in Section \ref{sec:bayeserr}. Theorem \ref{th:bayesianest} shows 
that the Bayesian error of the anatomized training data $D_A$ has smaller variance term than
the Bayesian error of the training data $D$. Hence, a model which is overfitted on the training data $D$ is likely
to be left out in the search space if the model is trained from the anatomized training data $D_A$.

The anatomized 1-NN has significantly lower error rate than the anonymized 1-NN at
the confidence intervals 0.99 and 0.98 when $k=l=2$, and
at the confidence interval 0.99 when $k=l=3$. The results 
aren't statistically significant for confidence intervals 
smaller than 0.95 or 0.99, as the anonymized 1-NN consistently doesn't fit one fold's training 
data. Its high error rate results in a significant increase in sample variance, reducing 
the statistical confidence. When we analyzed this training data, we noticed that the instance 
values were generalized to 
the root values of the generalization hierarchies which could eliminate the decision boundary in the original 
data. This observation emphasizes the anatomy's advantage for keeping the original attribute 
values despite diversifying the sensitive attribute values within a group. 

\subsection{Anatomized $k$-NN vs. Original $k$-NN}

\begin{figure*}[t!]
                    \centering
                    \begin{subfigure}[b]{0.45\textwidth}
                        \centering
                        \includegraphics[width=1\textwidth,height=0.8\textwidth]{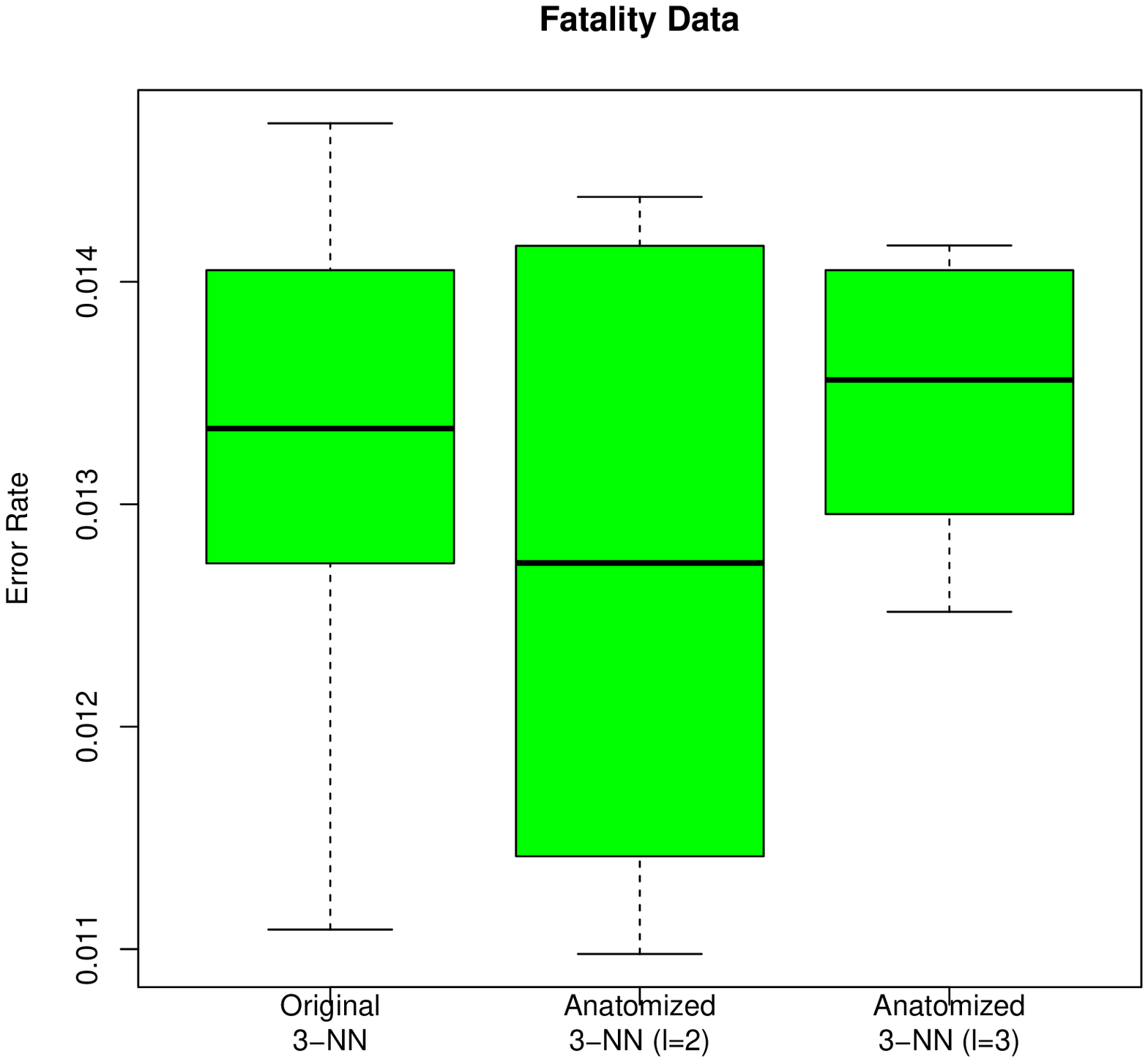}
                        \caption{3-NN}    
                        \label{fig:d0}
                    \end{subfigure}
                    \begin{subfigure}[b]{0.45\textwidth}  
                        \centering 
                        \includegraphics[width=1\textwidth,height=0.8\textwidth]{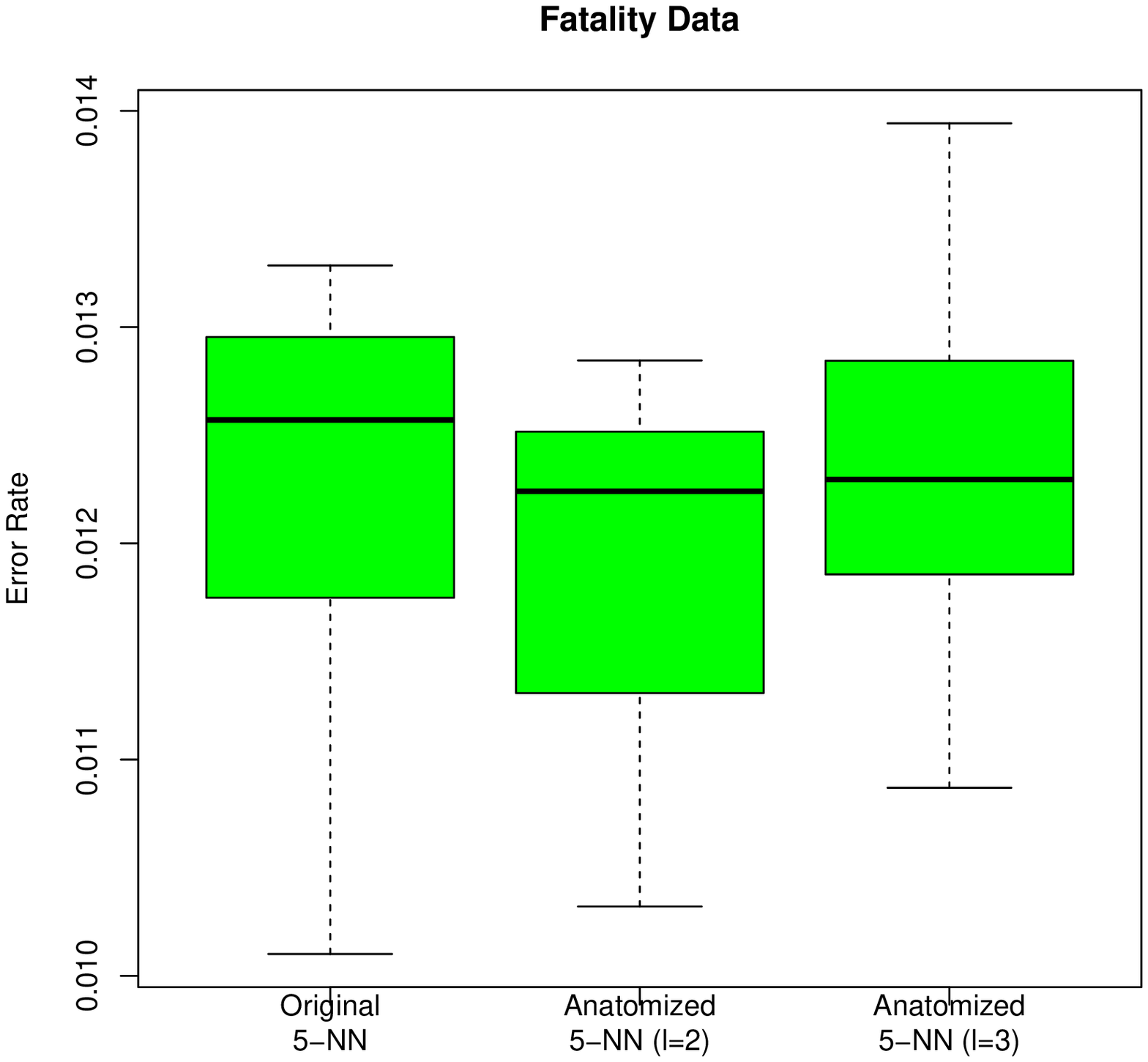}
                        \caption{5-NN}  
                        \label{fig:d2}
                    \end{subfigure}
                    \centering
                    \begin{subfigure}[b]{0.45\textwidth}
                        \centering
                        \includegraphics[width=1\textwidth,height=0.8\textwidth]{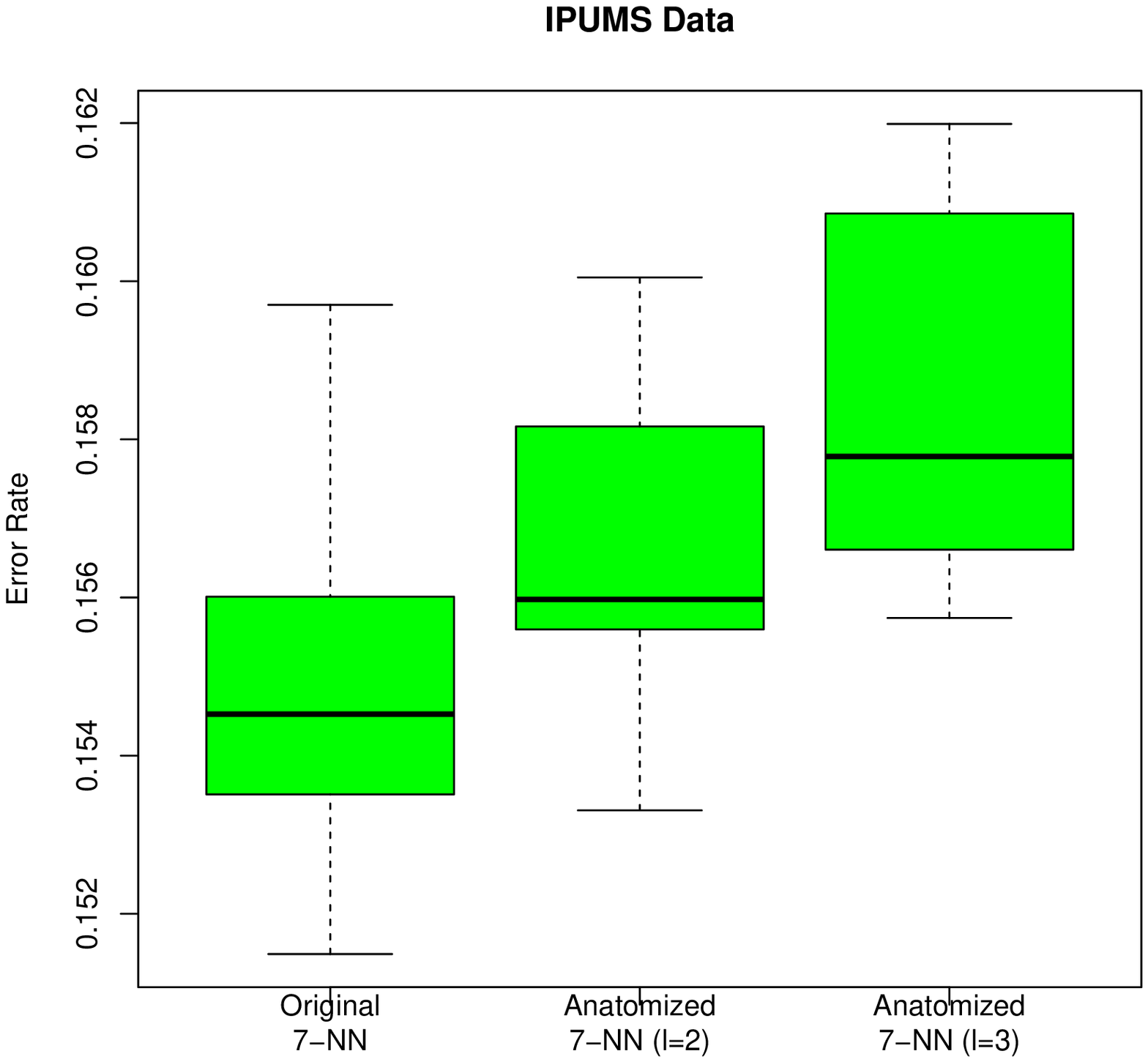}
                        \caption{7-NN}    
                        \label{fig:d0}
                    \end{subfigure}
                    \begin{subfigure}[b]{0.45\textwidth}  
                        \centering 
                        \includegraphics[width=1\textwidth,height=0.8\textwidth]{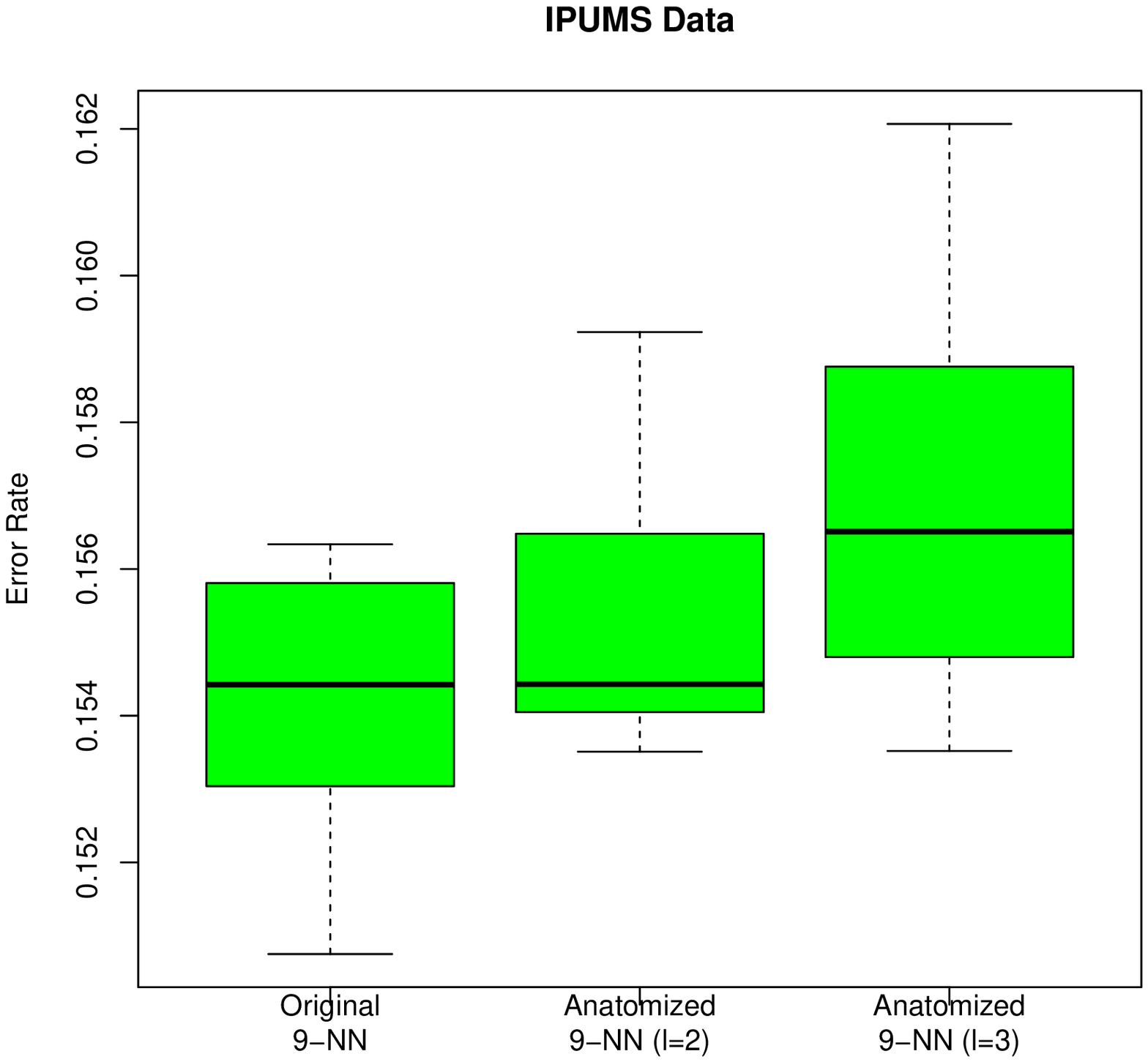}
                        \caption{9-NN}  
                        \label{fig:d2}
                    \end{subfigure}
                    \caption{Error Rates of $k$-NN Classifier vs Anatomized $k$-NN Classifier ($l=2$, $l=3$) on 10 Fold Cross Validation} 
                    \label{fig:errother}
\end{figure*}

In this section, we compare the anatomized $k$-NN classifier with the original $k$-NN classifier.
The comparison doesn't include the anonymized $k$-NN classifier because Inan et al.'s work considers 
only the anonymized 1-NN classifier \cite{kAnonSvm}. Its extension to $k>1$ cases is beyond the scope of this work.
Although we ran the experiments for anatomized 3-NN, 5-NN, 7-NN and 9-NN classifiers on the Adult, Bank
Marketing, Fatality and IPUMS datasets, we give the results on the larger Fatality 
and IPUMS datasets due to space limitations. We again include the cases of $l=2$ and $l=3$. Figure \ref{fig:errother} plots the error rate 
distributions of 3-NN and 5-NN classifiers on Fatality dataset, and 7-NN and 9-NN classifiers on IPUMS data. 

In the Fatality data, the anatomized 3-NN and 5-NN classifiers outperform the original 3-NN and 5-NN classifiers 
at the confidence intervals 0.99 and 0.98 when $l=2$. 
The anatomized 5-NN classifier also outperforms the
original 5-NN classifier at the confidence interval 0.95 when $l=2$.
In contrast, the original 3-NN and 5-NN classifiers outperform the anatomized 3-NN and 5-NN classifiers when $l=3$, although not to a statistically significant level.
For 3-NN classifiers, the average error rates are 0.0128, 0.0135 and 0.0132 for $D_A$ with $l=2$, $D_A$ with $l=3$ and 
original data respectively. On the other hand, the average error rates of 5-NN classifier on $D_A$ with $l=2$, $D_A$ with 
$l=3$ and original data are 0.0119, 0.0122 and 0.0122 respectively.

In the IPUMS data, the original 7-NN classifier outperforms the anatomized 7-NN classifier at the confidence intervals
0.99, 0.98, 0.95, 0.9 when $l=2$ and $l=3$. On the other hand, the original 9-NN classifier outperforms the 
anatomized 9-NN classifiers at the confidence interval 0.99 when $l=2$ and $l=3$.
 For 7-NN classifiers, the average error rates are 0.1567, 0.1586 and
0.1549 for $D_A$ with $l=2$, $D_A$ with $l=3$ and original data respectively. The average 
error rates of 9-NN classifier 
on $D_A$ with $l=2$, $D_A$ with $l=3$ and original data are 0.1552, 0.1568 and 0.1542 respectively.

In conclusion, the anatomized and original $k$-NN classifiers have similar statistically significant 
error rates for multiple values of $l$. 
These results confirm the theoretical analysis that we made in the earlier sections.

\begin{figure*}[htb!]
                    \centering
                    \begin{subfigure}[b]{0.4\textwidth}
                        \centering
                        \includegraphics[width=1\textwidth]{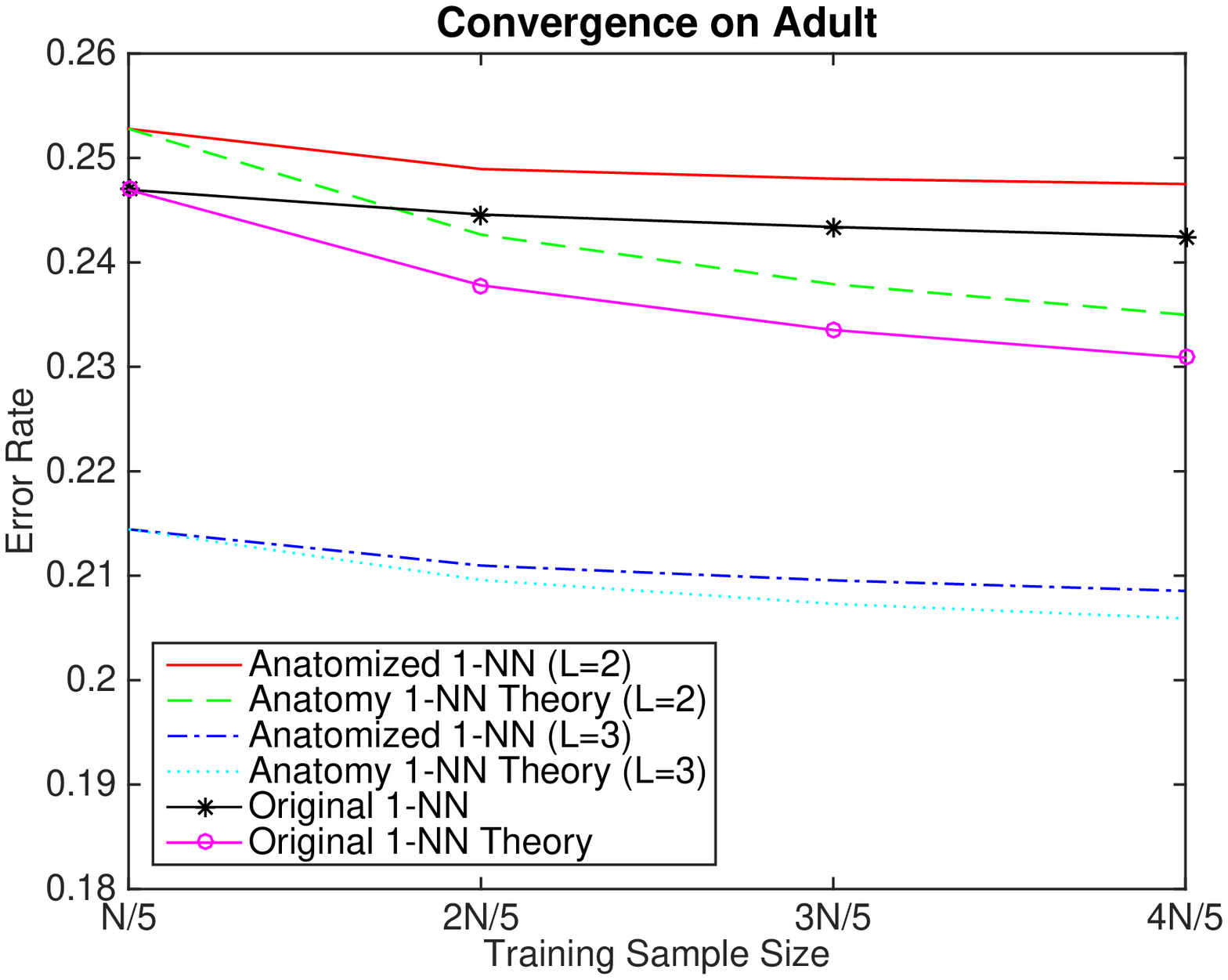}
                        \caption{Adult Data Error Rates}    
                        \label{fig:d0}
                    \end{subfigure}
                    \begin{subfigure}[b]{0.4\textwidth}  
                        \centering 
                        \includegraphics[width=1\textwidth]{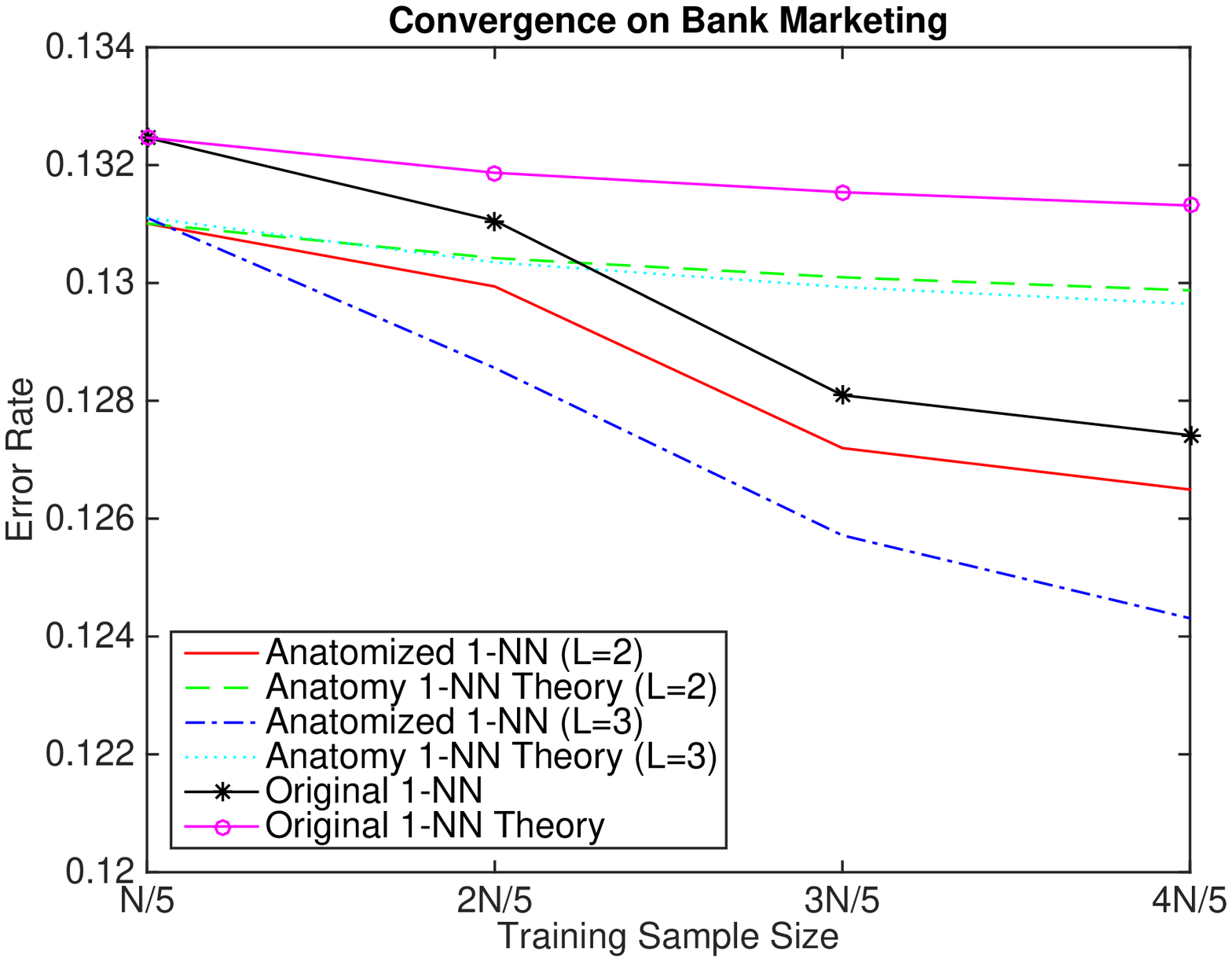}
                        \caption{Bank Marketing Data Error Rates}  
                        \label{fig:d2}
                    \end{subfigure}
                    \centering
                    \begin{subfigure}[b]{0.4\textwidth}
                        \centering
                        \includegraphics[width=1\textwidth]{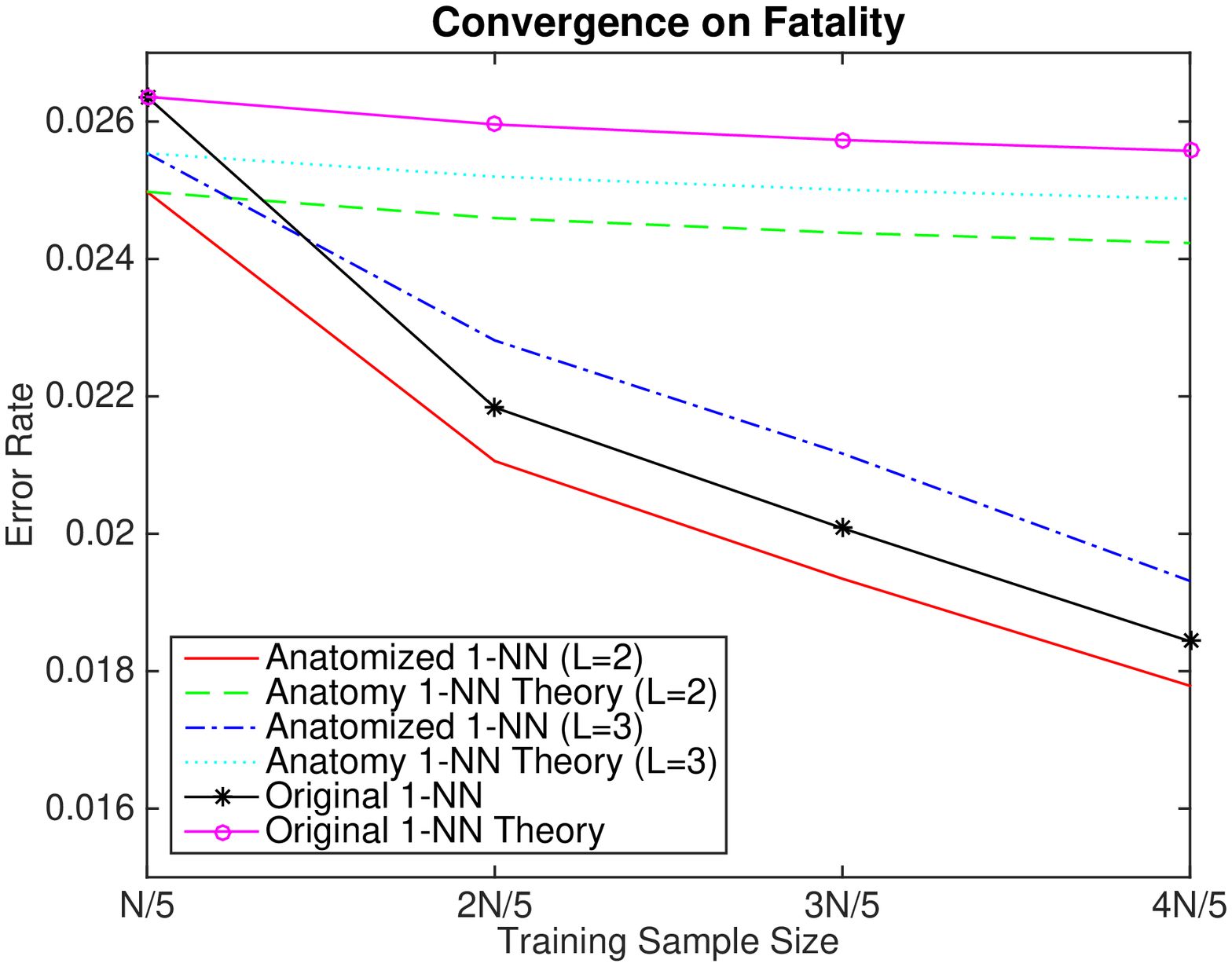}
                        \caption{Fatality Data Error Rates}    
                        \label{fig:d0}
                    \end{subfigure}
                    \begin{subfigure}[b]{0.4\textwidth}  
                        \centering 
                        \includegraphics[width=1\textwidth]{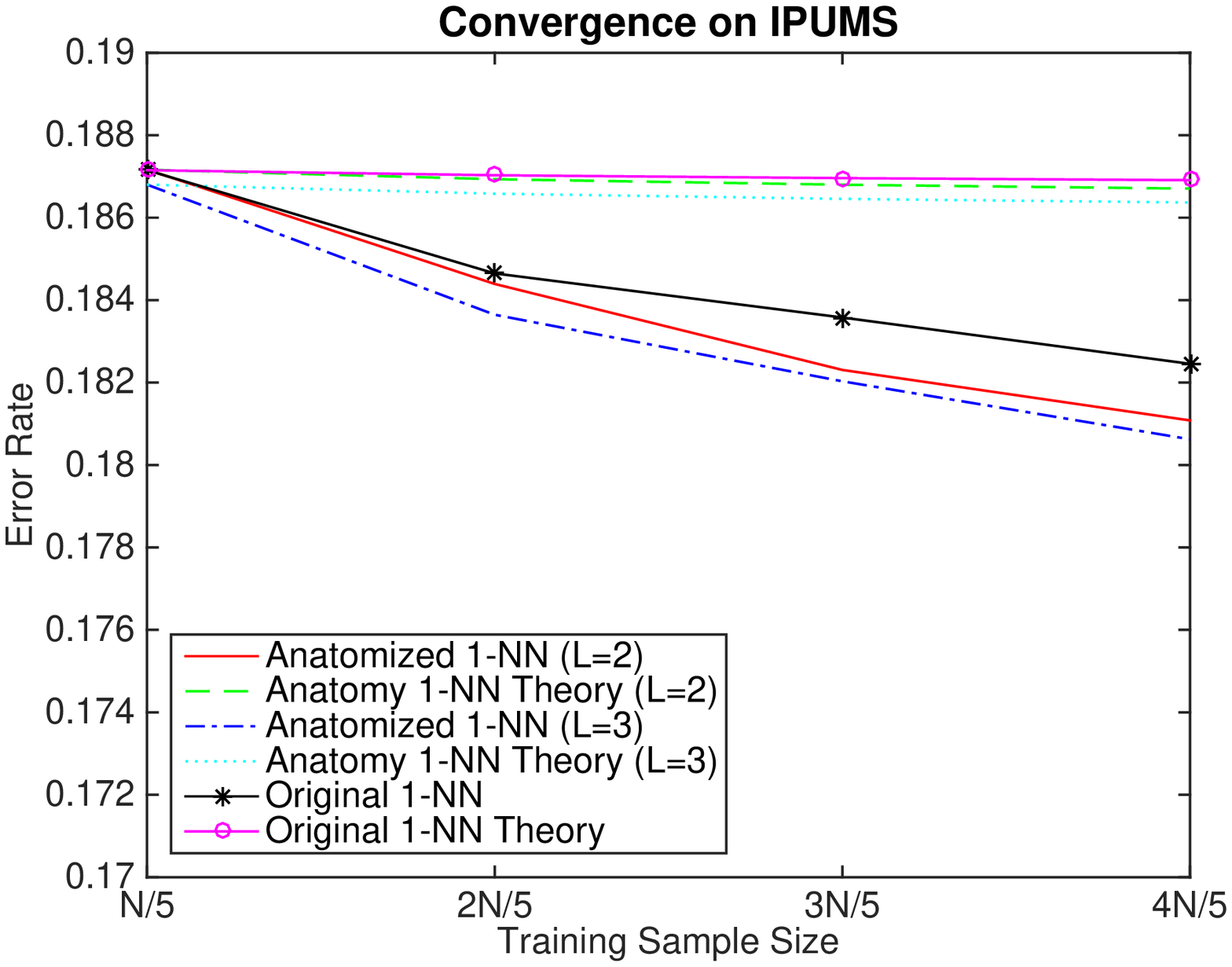}
                        \caption{IPUMS Data Error Rates}  
                        \label{fig:d2}
                    \end{subfigure}
                    \caption{Convergence Behavior of Original 1-NN Classifier vs Anatomized 1-NN Classifier ($l=2$, $l=3$)} 
                    \label{fig:conv}
\end{figure*}

\subsection{Convergence Behavior}

 We now compare the anatomized 1-NN classifier versus the original
1-NN classifier on convergence behavior. We create 5 partitions from
the Adult (after preprocessing), Bank Marketing, Fatality and IPUMS
datasets. Each partition is used as test data, and the
remaining 4 partitions are used incrementally for training.
Our objective is to show how the parameter $l$ in anatomized training data 
change the error rates when the training data size is increased incrementally. 
Figure  \ref{fig:conv} plots the average error 
rates for the original training data, the anatomized training data with $l=2$, 
the anatomized training data with $l=3$; and the theoretical error rate
in function of the training data size.

We can't know the asymptotical $R_A$ practically for theoretical error rates. 
We thus make the following estimation for the theoretical result. For each dataset, we 
set the $R_A$ to the minimum of the error rates in the specific dataset's results.
We then calculate the rate $\frac{1}{(Nl)^\frac{2}{d+1}}$
from the $N$, $d$ and $l$ values that we set in the experiments. Using the $R_A$ and
$\frac{1}{(Nl)^\frac{2}{d+1}}$, we computed the respective bias and eventually the
theoretical error rate according to the respective training data size and $l$.

The measured error rates in Figure \ref{fig:conv} show a convergence that is similar to 
the one that theoretical error rates show. Given the largest training data size $\frac{4N}{5}$;
0.015, 0.004, 0.008 and 0.0085 are approximately 
the maximum deviations of measured error rates from the theoretical error rates for the Adult, 
Bank Marketing, Fatality and the IPUMS datasets respectively.
We can also see that the convergence of error rate does not make 
much difference between the original data, anatomized data with $l=2$ and
the anatomized data with $l=3$. In all types of training data, the convergence
rate of 1-NN classifier is slow. 

\section{Conclusion}
\label{sec:conc}

This work demonstrates the feasibility of $k$-NN classification
using training data protected by anatomization under $l$-diversity.
We show that the asymptotic error bounds are the same for anatomized
data as for the original data.
Perhaps surprisingly, the proposed 1-NN classifier
has a faster convergence to the asymptotical error rate than the convergence of 1-NN
classifier using the training data without anatomization.
In addition, the analysis suggests that the
Bayesian error estimation for any non-parametric classifier using the anatomized training 
data reduces the variance term of the Bayesian 
error estimation, although it is hard to define the characteristic
of the bias term. 

Experiments on multiple datasets confirm the
theoretical convergence rates. These experiments also demonstrate
that proposed $k$-NN on anatomized data approaches or even outperforms
$k$-NN on original data. In particular, the experiments on well known Adult data show
that 1-NN on anatomized data outperforms learning on data anonymized
to the same anonymity levels using generalization.


\ifCLASSOPTIONcompsoc
  \section*{Acknowledgments}

\else
  \section*{Acknowledgment}
\fi
\label{sec:ack}

This work is supported by the 
``Anonymous''.
We thank 
``Anonymous''
for sharing his/her implementation used for evaluating
1-NN on generalization-based anonymization.
We also thank 
``Anonymous''
for helpful comments throughout
the theoretical analysis.



%
%
%
\bibliographystyle{IEEEtranS}
\bibliography{IEEEabrv,sigproc}

\end{document}